\documentclass{article}
\usepackage{iclr2022_workshop,times}

\usepackage{amsmath,amsfonts,bm}

\def\eqref#1{equation~\ref{#1}}
\def\1{\bm{1}}

\DeclareMathAlphabet{\mathsfit}{\encodingdefault}{\sfdefault}{m}{sl}
\SetMathAlphabet{\mathsfit}{bold}{\encodingdefault}{\sfdefault}{bx}{n}

\newcommand{\R}{\mathbb{R}}

\usepackage{amsmath}
\usepackage{amsfonts} 
\usepackage{array}

\newcommand{\model}{REMuS-GNN}
\newcommand{\NS}{Navier-Stokes equations}

\newcolumntype{Y}{>{\centering\arraybackslash}X}

\newcommand{\D}{\mathcal{D}}
\newcommand{\RR}{\mathcal{R}}
\newcommand{\Nin}{\mathcal{N}^{-}}

\usepackage{hyperref}
\usepackage{url}
\usepackage{subfig}
\usepackage{graphicx}
\usepackage{amsmath}
\usepackage{amssymb}
\usepackage{array}
\usepackage{tabularx}
\usepackage{wrapfig}
\usepackage[export]{adjustbox}
\usepackage{booktabs}
\usepackage{float}
\usepackage{mathtools}
\usepackage{multirow}
\usepackage{placeins}
\usepackage{multicol}

\title{REMuS-GNN: A Rotation-Equivariant Model
for Simulating Continuum Dynamics}

\author{
  Mario Lino \\
  Department of Aeronautics\\
  Imperial College London\\
  \texttt{mal1218@ic.ac.uk} \\
  \And
  Stathi Fotiadis \\
  Department of Bioengineering  \\
  Imperial College London \\
  \AND
  Anil A. Bharath \\
  Department of Bioengineering  \\
  Imperial College London \\
  \And
  Chris Cantwell \\
  Department of Aeronautics \\
  Imperial College London \\
}

\iclrfinalcopy % Uncomment for camera-ready version, but NOT for submission.
\begin{document}

\maketitle

\begin{abstract}
Numerical simulation is an essential tool in many areas of science and engineering, but its performance often limits application in practice or when used to explore large parameter spaces.
On the other hand, surrogate deep learning models, while accelerating simulations, often exhibit poor accuracy and ability to generalise.
In order to improve these two factors, we introduce \model{}, a rotation-equivariant multi-scale model for simulating continuum dynamical systems encompassing a range of length scales.
\model{} is designed to predict an output vector field from an input vector field on a physical domain discretised into an unstructured set of nodes.
Equivariance to rotations of the domain is a desirable inductive bias that allows the network to learn the underlying physics more efficiently, leading to improved accuracy and generalisation compared with similar architectures that lack such symmetry.
We demonstrate and evaluate this method on the incompressible flow around elliptical cylinders.

\end{abstract}

\section{Introduction}

Continuum dynamics models often describe the underlying physical laws by one or more partial differential equations (PDEs).
Numerical methods are well-established for approximately solving PDEs with high accuracy, however, they are computationally expensive \citep{spencer}.
Deep learning techniques have been shown to accelerate physical simulations \citep{Guo2016}, however, their relatively poor accuracy and limited ability to generalise restricts their application in practice.
Most deep learning models for simulating continuum physics have been developed around convolutional neural networks (CNNs) \citep{Thuerey2018,Wiewel2019}.
% In part, the success of CNNs lies in their translation invariance and locality \citep{goodfellow2016deep}, which represent desirable inductive biases for continuum-dynamics models.
CNNs constrain input and output fields to be defined on rectangular domains represented by regular grids, which is not suitable for more complex domains.
This has motivated the recent interest in graph neural networks (GNN) for learning to simulate continuum dynamics, which allow complex domains to be represented and the resolution spatially varied \citep{pfaff2020learning}.

Most physical processes possess several symmetries, among which translation and rotation are perhaps the most common.
Translation invariance/equivariance can be easily satisfied by using CNNs or GNNs \citep{battaglia2018relational}. For rotation, the general approach is to approximately learn invariance/equivariance by training against augmented training data which includes rotations \citep{chu2017data,li2018learning,mrowca2018flexible,lino2021simulating}.
Here, we propose \model{}, a \textit{R}otation \textit{E}quivariant \textit{Mu}lti-\textit{S}cale \textit{GNN} model that enforces the rotation equivariance of input and output vector fields, and improves the accuracy and generalisation over the data-augmentation approach.
\model{} forecasts the spatio-temporal evolution of continuum systems, discretised on unstructured sets of nodes, and processes the physical data at different resolutions or length scales, enabling the network to more accurately and efficiently learn spatially global dynamics.
We introduce a general directional message-passing (MP) algorithm and an edge-unpooling algorithm that is rotation invariant.
The accuracy and generalisation of \model{} was assessed on simulations of the incompressible flow around elliptical cylinders, an example of Eulerian dynamics with a global behaviour due the infinite speed of the pressure waves.

\section{Related work}
\paragraph{Neural solvers} During the last five years, most of the neural network models used for simulating continuum physics have included convolutional layers.
For instance, CNNs have been used to solve the Poisson's equation \citep{Tang2018,Ozbay2019} and to solve the \NS{} \citep{Guo2016,Thuerey2018,Lee2018,Kim2019,Wiewel2019}, achieving an speed-up of one to four orders of magnitude with respect to numerical solvers.
To the best of our knowledge \cite{alet2019graph} were the first to explore the use of GNNs to infer continuum physics by solving the Poisson PDE, and subsequently \cite{pfaff2020learning} proposed a mesh-based GNN to simulate a wide range of continuum dynamics.
Multi-resolution graph models were later introduced by \cite{li2020multipole,lino2021simulating,liu2021multi} and \cite{chen2021graph}.

\paragraph{Rotation invariance and equivariance}
There has been a constant interest to develop neural networks that are equivariant to symmetries, and particularly to rotation.
For instance, \cite{weiler2019general} and \cite{weiler20183d} introduced an SE(2)-equivariant and an SE(3)-equivariant CNNs respectively, both later applied to simulate continuum dynamics by \cite{wang2020incorporating} and \cite{siddani2021rotational}.
However, rotation equivariant CNNs only achieve equivariance with respect to a small set of rotations, unlike rotation equivariant GNNs \citep{thomas2018tensor,fuchs2020se,schutt2021equivariant,satorras2021n}.
All these rotation equivariant networks require a careful design of each of their components.
On the other hand, rotation invariant GNNs, such as directional MP neural networks \citep{klicpera2020directional,klicpera2021directional}, ensure the rotation invariance by a proper selection of the input and output attributes \citep{klicpera2021gemnet,liu2021spherical}.
Nevertheless, these GNNs can only be applied to scalar fields, but not vector fields \citep{klicpera2020directional}.
On the other hand, although Tensor Flow Networks \citep{thomas2018tensor} and SE(3)-Transformers \citep{fuchs2020se} can handle equivariant vector fields and could be applied to Eulerian dynamics, they lack an efficient mechanism for processing and propagating node features across long distances.
% \model{} allows vector fields as input and output, and it incorporates a multi-scale architecture. 

\section{\model{}}
For a PDE $\frac{\partial \bm{u}}{\partial t}=\mathcal{F}(\bm{u})$ on a spatial domain $\mathcal{D} \subset \R^2$, \model \ infers the temporal evolution of the two-dimensional vector field $\bm{u}(t,\bm{x})$ at a finite set of nodes $V^1$, with coordinates $\bm{x}^1_i  \in \mathcal{D}$.
Given an input vector field $\bm{u}(t_0, \bm{x}_{V^1})$ at time $t=t_0$ and at the $V^1$ nodes, a single evaluation of
\model{} returns the output field $\bm{u}(t_0+dt, \bm{x}_{V^1})$, where $dt$ is a fixed time-step size.
\model{} is equivariant to rotations of $\D$, i.e., if a two-dimensional rotation $\RR: \bm{x} \rightarrow R\bm{x}$ with $R \in \R^{2\times2}$ is applied to $\bm{x}_{V^1}$ and $\bm{u}(t_0, \bm{x}_{V^1})$ then the output field is $R\bm{u}(t_0+dt, \bm{x}_{V^1})$.
Such rotation equivariance is achieved through the selection of input attributes that are agnostic to the orientation of the domain (but still contain information about the relative position of the nodes) and the design of a neural network that is invariant to rotations.
\model{} is applied to a data structure expanded from a directed graph.
We denote it as $H^1:=(V^1,E^1,A^1)$, where $E^1:=\{(i,j)| i,j \in V^1\}$ is a set of directed edges and $A^1:=\{(i,j,k)|(i,j),(j,k) \in E^1\}$ is a set of directed angles.
The edges in $E^1$ are obtained using a k-nearest neighbours (k-NN) algorithm that guarantees that each node has exactly $\kappa$ incoming edges.
% Hence, $|E^1| = \kappa|V^1|$ and $|A^1|=\kappa^2|V^1|$.
Unlike traditional GNNs, there are no input node-attributes to the network.
The input attributes at edge $(i,j)$ are $\bm{e}_{ij}: = [u_{ij}, p(\bm{x}_j), \Omega_j]$, where $p(\bm{x})$ can be any physical parameter and $\Omega_j=1$ on Dirichlet boundaries and $\Omega_j=0$ elsewhere. 
The edge attribute $u_{ij}$ is defined as
\begin{equation}
    u_{ij} := \hat{\bm{e}}_{ij} \cdot \bm{u}(t_0,\bm{x}_j),
    \label{eq:projection}
\end{equation}
where $\hat{\bm{e}}_{ij}:=(\bm{x}_j - \bm{x}_i)/||\bm{x}_j - \bm{x}_i||_2$;
that is, $u_{ij}$ is the projection of the input vector field at node $j$ along the direction of the incoming edge $(i,j)$ (see Figure \ref{fig:proj_aggr}a).
The input attributes at angle $(i,j,k)$ are $\bm{a}_{ijk} := [||\bm{x}_j - \bm{x}_i||_2, ||\bm{x}_k - \bm{x}_j||_2, \cos(\alpha_{ijk}), \sin(\alpha_{ijk})],$ where $\alpha_{ijk}:=\measuredangle (i,j)(j,k)$.
Since all the input attributes are independent of the chosen coordinate system, any function applied exclusively to them is invariant to both rotations and translations.

We denote as $u'_{ij} \in \R$ to the output at edge $(i,j)$ of a forward pass through the network, and it represents the projection of the output vector field at node $j$ along the direction of the incoming edge $(i,j)$.
In order to obtain the desired output vectors at each node, $\bm{u}(t_0+dt,\bm{x}_{V^{1}})$, from these scalar values; we solve the overdetermined system of equations (if $\kappa > 2$) given by
\begin{equation}
    [\hat{\bm{e}}^1_{1:\kappa,j}][\bm{u}(t_0+dt,\bm{x}_j)] = [u'_{1:\kappa,j}], \ \ \ \forall j \in V^1,
    \label{eq:aggr}
\end{equation}
where matrix $[\hat{\bm{e}}_{1:\kappa,j}] \in \R^{\kappa \times 2}$ contains in its rows the unit vectors along the directions of the $\kappa$ incoming edges at node $j$, $[\bm{u}(t_0+dt,\bm{x}_j)] \in \R^{2\times1}$ is a column vector with the horizontal and vertical components of the output vector field at node $j$, and $[u'_{1:\kappa,j}] \in \R^{\kappa \times 1}$ is another column vector with the value of $u'$ at each of the $\kappa$ incoming edges.
This step can be regarded as the inverse of the projection in equation (\ref{eq:projection}).
To solve equation (\ref{eq:aggr}) we use the Moore-Penrose pseudo-inverse of $[\hat{\bm{e}}_{1:\kappa,j}]$, which we denote as  $[\hat{\bm{e}}_{1:\kappa,j}]^{+} \in \R^{2 \times \kappa}$.
Thus, if we define the \textit{projection-aggregation} function at scale $\ell$, $\rho^\ell: \R^\kappa \rightarrow \R^2$, as the matrix-vector product given by
\begin{equation}
    \rho^\ell(e_1, e_2, \dots, e_\kappa) := [\hat{\bm{e}}^\ell_{1:\kappa,j}]^{+} [e_1, e_2, \dots, e_\kappa]^T
    \label{eq:solve_aggr}
\end{equation}
then $\bm{u}(t_0+dt,\bm{x}_j) = \rho^1(u'_{1:\kappa,j})$ (see Figure \ref{fig:proj_aggr}b).

The output $\bm{u}(t_0+dt,\bm{x}_{V^{1}})$ can be successively re-fed to \model{} to produce temporal roll-outs.
In each forward pass the information processing happens at $L$ length-scales in a U-Net fashion, as illustrated in Figure \ref{fig:net}.
A single MP layer applied to $H^1$ propagates the edge features only locally between adjacent edges.
To process the information at larger length-scales, \model{} creates an $H$ representation for each level, where MP is also applied.
The lower-resolution representations ($H^2, H^3, \dots, H^L$; with $|V_1| > |V_2| > \dots > |V_L|$) possess fewer nodes, and hence, a single MP layer can propagate the features of edges and angles over longer distances more efficiently.
Each $V^{\ell+1}$ is a subset of $V^{\ell}$ obtained using Guillard's coarsening algorithm \citep{guillard1993node}, 
and $E^{\ell+1}$ and $A^{\ell+1}$ are obtained in an analogous manner to how $E^{1}$ and $A^{1}$ were obtained, as well as their attributes $\bm{e}_{ij}^{\ell+1}$ and $\bm{a}_{ij}^{\ell+1}$.
Before being fed to the network, all the edge attributes $\bm{e}_{ij}^{\ell}$ and angle attributes $\bm{a}_{ij}^{\ell}$ are encoded through independent multi-layer perceptrons (MLPs).
At the end, another MLP decodes the output edge-features of $E^1$ to return $u'_{ij}$.
As depicted in Figure \ref{fig:net}, the building blocks of \model{}'s network are a directional MP (EdgeMP) layer, an edge-pooling layer and an edge-unpooling layer.

\paragraph{EdgeMP layer}
Based on the GNBlock introduced by \cite{Sanchez-Gonzalez2018} and \cite{battaglia2018relational} to update node and edge attributes, we define a general MP layer to update angle and edge attributes.
The angle-update, angle-aggregation and edge-update at scale $\ell$ are given by
\begin{alignat}{2}
        \bm{a}^{\ell}_{ijk} &\leftarrow f^{a}([\bm{a}^{\ell}_{ijk},\bm{e}^{\ell}_{ij},\bm{e}^{\ell}_{jk}]), \ \ \ &\forall (i,j,k) \in A^\ell,
        \label{eq:angle_model} \\
        \overline{\bm{a}}^{\ell}_{jk} &\leftarrow \frac{1}{\kappa} \sum_{k \in \Nin_j} \bm{a}^{\ell}_{ijk}, \ &\forall (j,k) \in E^\ell,
        \label{eq:angle_aggr} \\
        \bm{e}^{\ell}_{jk} &\leftarrow f^e([\bm{e}_{jk}, \overline{\bm{a}}^{\ell}_{jk}]), \ &\forall (j,k) \in E^\ell.
        \label{eq:edge_model}
\end{alignat}
Functions $f^a$ and $f^e$ are MLPs in the present work. This algorithm is illustrated in Figure \ref{fig:EdgeMP}.

\paragraph{Edge-pooling layer}
Given a node $j \in V^{\ell}$ (hence $j \in V^{\ell-1}$ too), an outgoing edge $(j,k) \in E^{\ell}$ and its $\kappa$ incoming edges $(i,j) \in E^{\ell-1}$, we can define $\kappa$ new angles $(i,j,k) \in A^{\ell-1,\ell}$ that connect scale $\ell-1$ to scale $\ell$.
% The angle set $A^{\ell,\ell+1}:=\{(i,j,k) | (i,j) \in E^{\ell}, (j,k) \in E^{\ell+1}\}$ is responsible for connecting scales $\ell$ and $\ell+1$, and it has $\kappa^2|V^{\ell+1}|$ elements.
% Their input attributes, $\bm{a}_{ijk}^{\ell,\ell+1}$, are obtained like $\bm{a}_{ijk}^{1}$ were.
Pooling from $H^{\ell-1}$ to $H^{\ell}$ is performed as the EdgeMP, but using the incoming edges $(i,j) \in E^{\ell-1}$, outcoming edge $(j,k) \in E^{\ell}$ and angles $(i,j,k) \in A^{\ell-1,\ell}$.

\paragraph{Edge-unpooling layer}
To perform the unpooling from $H^{\ell+1}$ to $H^{\ell}$ we first aggregate the features of incoming edges into node features.
Namely, given the $\kappa$ incoming edges at node $j \in V^{\ell+1}$ and their $F$-dimensional edge-features, $\bm{e}^{\ell+1}_{ij} = [(e_1)^{\ell+1}_{i,j}, (e_2)^{\ell+1}_{i,j}, \dots, (e_F)^{\ell+1}_{i,j}]$, the node-feature matrix $W_j^{\ell+1} \in R^{2\times F}$ is obtained applying the projection-aggregation function $\rho^{\ell+1}$ to each component of the edge features according to 
\begin{equation}
    W_j^{\ell+1} = \bigg[
    \rho^\ell \big( (e_1)^{\ell+1}_{1:\kappa,j} \big)^T
    \Big|
    \rho^\ell \big( (e_2)^{\ell+1}_{1:\kappa,j} \big)^T
    \Big| \dots \Big|
    \rho^\ell \big( (e_F)^{\ell+1}_{1:\kappa,j} \big)^T
    \bigg],
    \label{eq:desproject_unpool}
\end{equation}
where $\cdot \ | \ \cdot$ denotes the horizontal concatenation of column vectors.
Then, $W_j^{\ell+1}$ is interpolated to the set of nodes $V^{\ell}$ following the interpolation algorithm introduced by \cite{qi2017pointnet++}, yielding $W_k^{\ell} \in \R^{2\times F}$ at each node $k \in V^{\ell}$.
Next, these node features are projected to each edge $(l,k)$ on $E^{\ell}$ to obtain $\bm{w}^{\ell}_{lk} \in \R^{F}$, where
\begin{equation}
    \bm{w}^{\ell}_{lk} := \hat{\bm{e}}_{lk}^{\ell} W_k^{\ell} , \ \ \ \forall (l,k) \in E^{\ell}.
    \label{eq:projection_unpool}
\end{equation}
Finally, the MLP $f^u$ is used to update the edge features $\bm{e}^{\ell}_{lk}$ as
\begin{equation}
    \bm{e}^{\ell}_{lk} \leftarrow f^u([\bm{e}^{\ell}_{lk}, \bm{w}_{lk}^\ell]), \ \ \ \forall (l,k) \in E^{\ell}.
    \label{eq:update_unpool}
\end{equation}

\section{Experiments}

\paragraph{Datasets} Datasets \texttt{Ns} and \texttt{NsVal} where use for training and validation respectively.
Both contain solutions of the incompressible \NS{} for the flow around an elliptical cylinder with an aspect ratio $b \in [0.5,0.8]$ on a rectangular fluid domain with top and bottom boundaries separated by a distance $H \in [5,6]$.
% The free-stream velocity on the inlet, bottom and top boundaries is horizontal and with magnitude equal to one.
Each sample consists of the vector-valued velocity field at 100 time-points within the periodic vortex-shedding regime with a Reynolds number $Re \in [500,1000]$.
Domains are discretised with between 4800 and 9200 nodes.
In \model{}, the input $\bm{u}(t,\bm{x})$ corresponds to the velocity field and $p(\bm{x})$ to $Re$.
Besides these datasets, six datasets with out-of-distribution values for $b$, $H$ and $Re$ were used for assessing the generalisation of the models; and dataset \texttt{NsAoA} includes rotated ellipses.
Further details are included in Appendix \ref{sec:datasets_details}.

\paragraph{Models}
We compare \model{} with MultiScaleGNN \citep{lino2021simulating}, a state-of-the-art model for inferring continuum dynamics; and with 
MultiScaleGNNg, a modified version of MultiScaleGNN.
MultiScaleGNN and MultiScaleGNNg possess the same U-Net-like architecture as \model{}, while MultiScaleGNNg and \model{} also share the same low-resolution representations.
The benchmark models are not rotation equivariant, so they were trained with and without rotations of the domain.
All the models have three scales ($L=3$) and a similar number of learnable parameters.
For hyper-parameter choices and training setup see Appendix \ref{sec:models_details}.

\paragraph{Results}
\begin{wrapfigure}[18]{L}{0.5\columnwidth}
\begin{center}
\includegraphics[clip, width=0.48\columnwidth, trim={0mm, 0mm, 0mm, 0mm}]{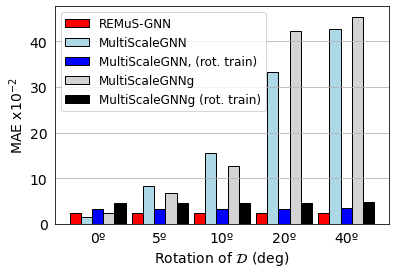}
\end{center}
\caption{\small MAE ($\times 10^{-2}$) on the \texttt{NsVal} dataset under rotations of $\D$.
\label{fig:rots_vs_mae}
}
\end{wrapfigure}
From Figure \ref{fig:rots_vs_mae} it is clear that the MAEs of the predictions of MultiScaleGNN and MultiScaleGNNg trained without rotations grow significantly during inference on rotations of the domain greater than 5 degrees.
On the other hand, these models trained with rotations are able to maintain an approximately constant MAE for each rotation of the domain.
Hence, from now on we only consider these benchmark models.
It can be seen that \model{} outperforms the benchmark models on every rotation of the validation dataset.
Table \ref{table:mae_val_test} collects the MAE on the velocity field (measured with the free-stream as reference) and the MAE on the $x$-coordinate of the separation point on the upper wall of the ellipses.
It can be concluded that \model{} also has a much better generalisation 
than the benchmark models.
Additional results are included in Appendix \ref{sec:results}.

\begin{table}[ht]
\centering
\caption{\small MAE $\times 10^{-2}$ in the velocity field and the $x$-coordinate of the separation point}
\label{table:mae_val_test}
\begin{center}
\footnotesize
\begin{tabularx}{\textwidth}{llYYYYYYY} 
\toprule
& & \texttt{\scriptsize NsLowRe} & \texttt{\scriptsize NsHighRe} & \texttt{\scriptsize NsThin} & \texttt{\scriptsize NsThick} & \texttt{\scriptsize NsNarrow} & \texttt{\scriptsize NsWide} & \texttt{\scriptsize NsAoA} \\
\midrule
\multirow{2}{*}{\scriptsize \model{}} &
\scriptsize{Velocity field} &
\textbf{4.514} & \textbf{9.576} & \textbf{3.152} & \textbf{4.430} & \textbf{2.861} & \textbf{2.873} & \textbf{2.504}\\
& \scriptsize{Separation point} &
\textbf{3.082} & \textbf{6.464} & \textbf{2.477} & \textbf{4.488} & \textbf{2.934} & \textbf{2.964} & \textbf{3.219}\\
\midrule
\multirow{2}{*}{\scriptsize MultiScaleGNN} &
\scriptsize{Velocity field}\scriptsize{} &
5.723 & 13.886 & 3.703 & 5.531 & 3.583 & 3.454 & 3.451 \\
& \scriptsize{Separation point} &
4.424 & 7.524 & 2.825 & 4.873 & 3.830 & 3.959 & 4.386\\
\midrule
\multirow{2}{*}{\scriptsize MultiScaleGNNg} &
\scriptsize{Velocity field} &
4.826 & 8.552 & 4.085 & 7.201 & 4.759 & 4.666 & 4.593\\
& \scriptsize{Separation point} &
4.414 & 7.264 & 3.025 & 6.405 & 4.222 & 4.468 & 4.993\\
\bottomrule
\end{tabularx}
\end{center}
\end{table}

\section{Conclusion}
We proposed a translation and rotational equivariant model for predicting the spatio-temporal evolution of vector fields defined on continuous domains, where the dynamics may encompass a range of length scales and/or be spatially global.
The proposed model employs a generalised directional message-passing algorithm and a novel edge-unpooling algorithm specifically designed to satisfy the rotation invariance requirement.
The incorporation of rotation equivariance as a strong inductive bias results in a higher accuracy and better generalisation compared with the vanilla data-augmentation approach for approximately learning the rotation equivariance.
To the best of the authors' knowledge, \model{} is the first multi-scale and rotation-equivariant GNN model for inferring Eulerian dynamics.

\FloatBarrier

\bibliography{iclr2022_workshop}
\bibliographystyle{iclr2022_workshop}

\appendix

\newpage

\section{Diagrams of \model{}'s building blocks}

\begin{figure}[ht]
\centering
\subfloat[]{\includegraphics[scale=0.25]{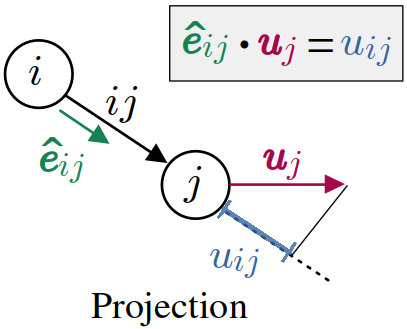}}
\quad
\quad
\quad
\subfloat[]{\includegraphics[scale=0.25]{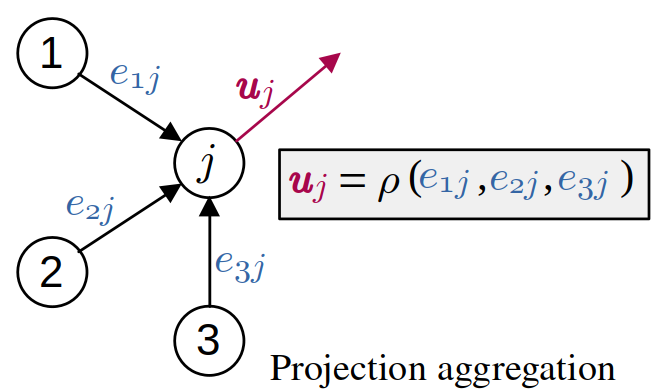}}
\caption{\small Diagrams of the projection step and projection-aggregation step (with $\kappa=3$).
The projection allows to encode a vector defined at node $j$, $\bm{u}_j$, as its projection along the direction of the $\kappa$ incoming edges.
The projection-aggregation is equivalent to the inverse of the projection step. Given the projection of $\bm{u}_j$ along the $\kappa$ incoming edges, $e_{1:\kappa,j}$; it restores the vector $\bm{u}_j$ by solving the overdetermined system of equations given by equation (\ref{eq:aggr}). 
\label{fig:proj_aggr}}
\end{figure}

\begin{figure}[ht]
\centering
\includegraphics[clip,width=0.85\columnwidth, trim={0mm 0mm 0mm 0mm}]{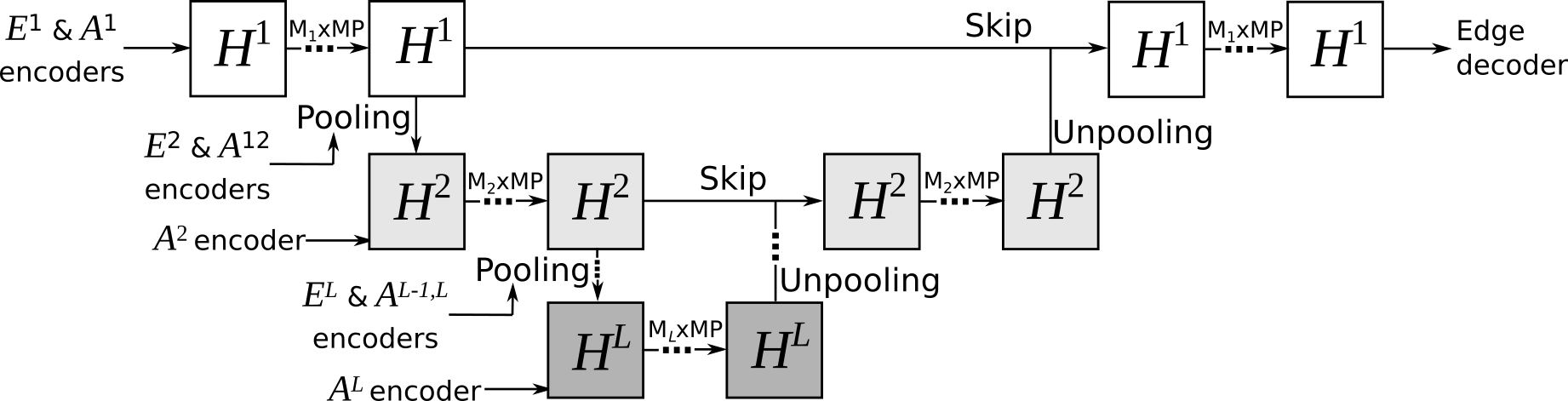}
\caption{\small \model{} architecture. $H^1$ is a high-resolution representation, $H^\ell$ with $\ell \geq 2$ are lower-resolution representations. 
\label{fig:net}
}
\end{figure}

\begin{figure}[ht]
\centering
\includegraphics[width=0.9\columnwidth]{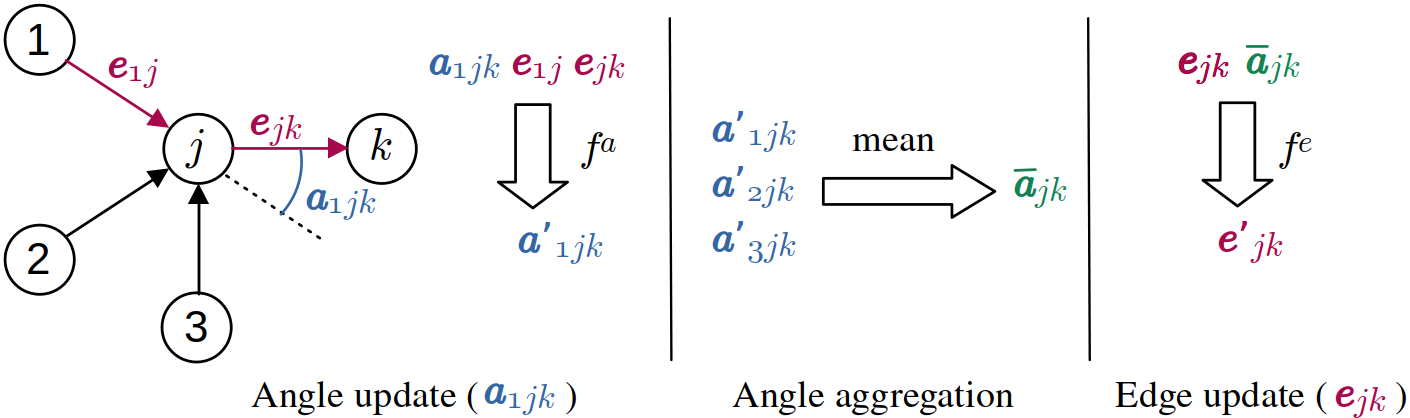}
\caption{\small Diagram of the EdgeMP algorithm applied to update the edge attribute $\bm{e}_{jk}$ (in the diagram $\kappa = 3$). The algorithm has three steps: update of the angle attributes, aggregation of the angle attributes and update of the edge attributes. In this case the angle-update step must be repeated 3 times (once for each incoming edge at node $j$).
\label{fig:EdgeMP}
}
\end{figure}

\FloatBarrier

\section{Rotation invariance of the edge-unpooling layer}
\label{sec:unpooling}
To achieve the rotation equivariance of \model{} it is required that the MP layers, edge-pooling layers and edge-unpooling layers are invariant to rotations. 
It is easy to see that MP and pooling layers (equations (\ref{eq:angle_model}) to (\ref{eq:edge_model})) are rotation invariant.
As for the unpooling from scale $\ell+1$ to scale $\ell$, the edge-unpooling layers in \model{} perform four steps:
\begin{enumerate}
    \item Aggregation of the incoming-edges' feature-vectors at each node in $V^{\ell+1}$ \textemdash equation (\ref{eq:desproject_unpool}).
    \item Interpolation of the obtained features from $V^{\ell+1}$ to $V^{\ell}$ \textemdash knn-interpolation in \citep{qi2017pointnet++}.
    \item Projection of the interpolated node-features along the direction of the incoming edges at scale $\ell$ \textemdash equation (\ref{eq:projection_unpool}).
    \item Update of the edge features \textemdash equation (\ref{eq:update_unpool}).
\end{enumerate}

Steps 2 and 4 are invariant to rotations since they do not depend on the particular directions of the edges on $V^{\ell}$ nor $V^{\ell+1}$.
On the other hand, step 1 is not rotation invariant since a two-dimensional rotation $\RR: \bm{x} \rightarrow \bm{x}R$ of $\D$ (and $V^1$) modifies the output of
the projection-aggregation function given by equation (\ref{eq:solve_aggr}) to
\begin{alignat*}{1}
    ([\hat{\bm{e}}^\ell_{1:\kappa,j}]R)^{+} [e_1, e_2, \dots, e_\kappa]^T &= 
    R^{+}[\hat{\bm{e}}^\ell_{1:\kappa,j}]^{+} [e_1, e_2, \dots, e_\kappa]^T \\ &= 
    R^{-1}\rho^{\ell+1}(e_1, e_2, \dots, e_\kappa).
\end{alignat*}
Hence, according to equation (\ref{eq:desproject_unpool}), the result of the step 1 in the edge-unpooling layer is $R^{-1}W_j^{\ell+1}$ for all $j \in V^{\ell+1}$, and the result of step 2 (rotation invariant) is $R^{-1}W_k^{\ell}$ for all $k \in V^{\ell}$. 
Step 3 is not rotation invariant, and given the input $R^{-1}W_k^{\ell}$, the output that follows is
\begin{alignat*}{1}
    (\hat{\bm{e}}_{lk}^{\ell}R) (R^{-1}W_k^{\ell}) &= \hat{\bm{e}}_{lk}^{\ell}W_k^{\ell} \\ &= \bm{w}^{\ell}_{lk}.
\end{alignat*}
Thus, despite step 1 and 3 are not invariant to rotations separately, they are when applied jointly.

\section{Datasets details}
\label{sec:datasets_details}

\begin{wrapfigure}[9]{R}{0.35\columnwidth}
\begin{center}
\includegraphics[clip, width=0.35\columnwidth, trim={0mm, 0mm, 0mm, 0mm}]{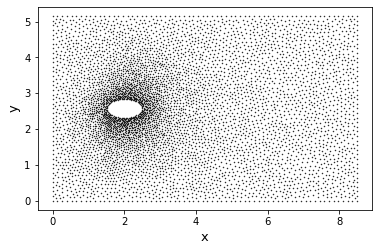}
\end{center}
\caption{\small Set of nodes $V_1$ for a simulation on the \texttt{Ns} dataset.
\label{fig:nodes}
}
\end{wrapfigure}

We solved the two-dimensional incompressible Navier-Stokes equation using the high-order solver Nektar++ \citep{nektar}.
The Navier-Stokes equations read
\begin{gather}
        \frac{\partial u}{\partial x} + \frac{\partial v}{\partial y} = 0, \\
        \frac{\partial u}{\partial t} + u \frac{\partial u}{\partial x} + v\frac{\partial u}{\partial y} = - \frac{\partial p}{\partial x} + \frac{1}{Re}\bigg( \frac{\partial^2 u}{\partial x^2} + \frac{\partial^2 u}{\partial y^2} \bigg), \\
        \frac{\partial v}{\partial t} + u \frac{\partial v}{\partial x} + v\frac{\partial v}{\partial y} = - \frac{\partial p}{\partial y} + \frac{1}{Re}\bigg( \frac{\partial^2 v}{\partial x^2} + \frac{\partial^2 v}{\partial y^2} \bigg), 
\label{eq:ns}
\end{gather}
where $u(t,x,y)$ and $v(t,x,y)$ are the horizontal and vertical components of the velocity field, $p(t,x,y)$ is the pressure field, and $Re$ is the Reynolds number.
We consider the flow around an elliptical cylinder with major axis equal to one and minor axis $b$ on a rectangular fluid domain of dimensions $8.5 \times D$ (see Figure \ref{fig:nodes}).
The left, top and bottom boundaries have as boundary condition $u = 1$, $v=0$ and $\partial p/ \partial x = 0$; the right boundary is an outlet with $\partial u/ \partial x = 0$, $\partial v/ \partial x=0$ and $p = 0$; and the cylinder wall has a no-slip condition $u=v=0$. 
In our simulations we only selected $Re$ values that yield solutions within the laminar vortex-shedding regime.
The sets of nodes $V^1$ employed for each simulation were created using \textit{Gmsh} with an element-size equal to $h$ at the corners of the domain and $0.3h$ on the cylinder wall.
Each simulation contains $100$ time-points equispaced by a time-step size $dt=0.1$.
The parameters of the training, validation and testing datasets are collated in Table \ref{table:datasets}.

\begin{table}[ht]
\centering
\caption{\small Incompressible flow datasets}
\label{table:datasets}
\footnotesize
\begin{tabular}{lccccccc} 
\toprule
Dataset                 & $Re$ & $b$ & $H$ & AoA (deg) & $h$ & \#Simulations    & Purpose   \\ 
\midrule
\texttt{Ns}       & 500-1000  & 0.5-0.8 & 5-6 & 0    & 0.10-0.16 & 5000 & Training \\
\texttt{NsVal}    & 500-1000  & 0.5-0.8 & 5-6 & 0    & 0.10-0.16 & 500  & Validation \\
\texttt{NsLowRe}  & 200-500  & 0.5-0.8 & 5-6 & 0    & 0.10-0.16 & 500  & Testing \\
\texttt{NsHighRe} & 1000-1500 & 0.5-0.8 & 5-6 & 0    & 0.10-0.16 & 500  & Testing \\
\texttt{NsThin}   & 500-1000  & 0.3-0.5 & 5-6 & 0    & 0.10-0.16 & 500  & Testing \\
\texttt{NsThick}  & 500-1000  & 0.8-1.0 & 5-6 & 0    & 0.10-0.16 & 500  & Testing \\
\texttt{NsNarrow} & 500-1000  & 0.5-0.8 & 4-5 & 0    & 0.10-0.16 & 500  & Testing \\
\texttt{NsWide}   & 500-1000  & 0.5-0.8 & 6-7 & 0    & 0.10-0.16 & 500  & Testing \\
\texttt{NsAoA}    & 500-1000  & 0.5-0.8 & 5.5 & 0-10 & 0.12 & 240  & Testing \\
\bottomrule
\end{tabular}
\end{table}

\section{Models details} \label{sec:models_details}
The implementation of the benchmark model MultiScaleGNN is taken from \cite{lino2021simulating}.
MultiScaleGNNg is a modified version of MultiScaleGNN to follow the pooling and unpooling used by \cite{liu2021multi}.
For this model, the low-resolution sets of nodes were generated using Guillard's coarsening algorithm \citep{guillard1993node} as in \model{}.
This way, both models share the same high and low-resolution discretisations.
For a fair comparison all the models considered in the present work follow the same U-Net-like architecture and have in common the hyper-parameters and training setup described below.
Hence, they have a similar number of learnable parameters ($\sim 2.2$M).

\paragraph{Hyper-parameters choice}
The number of incoming edges at each node was set to $\kappa=5$, and the number of linear layers in each MLP is 2 (except for $f^u$, which has 3 linear layers), with 128 neurons per hidden layer.
All MLPs use SELU activation functions \citep{klambauer2017self}, and, batch normalisation \citep{ba2016layer}.
The number of MP layers we used at each scale are $2\times M_1 = 8$, $2 \times M_2 = 4$ and $M_3 = 4$.

\paragraph{Training details}
During training four graphs were fed per iteration.
First, each training iteration predicted a single time-point, and every time the training loss decreased below 0.02 we increased the number of iterative time-steps by one, up to a limit of 10.
We used the loss function given by 
\begin{align}
    \mathcal{L} = \mathrm{MSE}&\Big(\hat{\mathbf{u}}(t, \mathbf{x}^1_{V^1}),\mathbf{u}(t, \mathbf{x}^1_{V^1}) \Big) \nonumber\\
    &+ \lambda_d\, \mathrm{MAE}\Big(\hat{\mathbf{u}}(t, {\mathbf{x}^1_{V^1} \in \partial_D\mathcal{D}}),\mathbf{u}(t, \mathbf{x}^1_{V^1} \in \partial_D\mathcal{D}) \Big) \nonumber,
\label{eq:loss}
\end{align}
with $\lambda_d=0.25$.
The initial time-point $t_0$ was randomly selected for each prediction, and, we added to the initial field noise following a uniform distribution between -0.01 and 0.01. 
After each time-step, the models' weights were updated using the Adam optimiser with its standard parameters \citep{kingma2014adam}.
The learning rate was set to $10^{-4}$ and multiplied by 0.5 when the training loss did not decrease after two consecutive epochs, also, we applied gradient clipping to keep the Frobenius norm of the weights' gradients below or equal to one.

\section{Additional results}
\label{sec:results}
\begin{figure}[ht]
\centering
\begin{tabular}{ccc}
\includegraphics[clip, width=0.3\textwidth, trim={15mm 15mm 25mm 5mm} ]{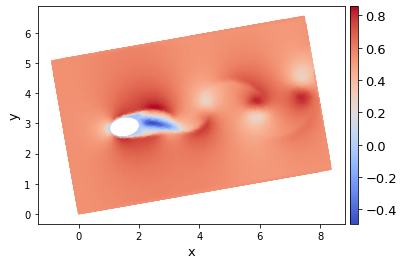} &
\includegraphics[clip, width=0.3\textwidth, trim={15mm 15mm 25mm 5mm}]{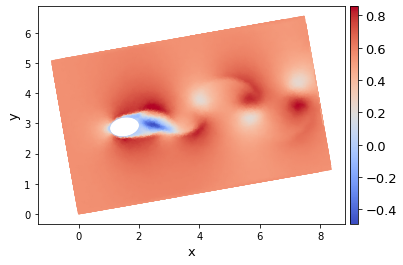} &
\includegraphics[clip, width=0.3\textwidth, trim={15mm 15mm 25mm 5mm}]{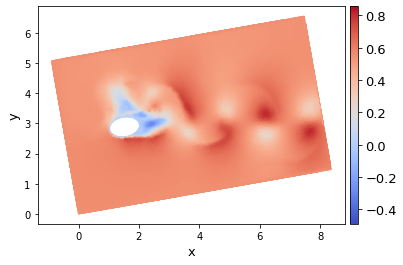}
\\
(a) Target & (b) \model{} & (c) MultiScaleGNN 
\\
 & (MAE = 0.0210) & train no rots.  (MAE = 0.1051)
\\
\includegraphics[clip, width=0.3\textwidth, trim={15mm 15mm 25mm 5mm}]{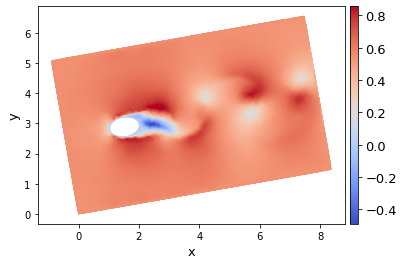} &
\includegraphics[clip, width=0.3\textwidth, trim={15mm 15mm 25mm 5mm}]{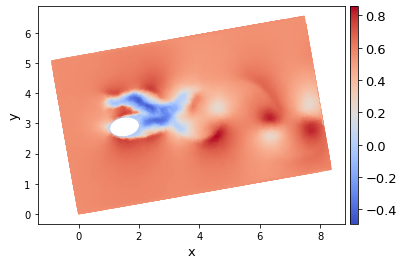} &
\includegraphics[clip, width=0.3\textwidth, trim={15mm 15mm 25mm 5mm}]{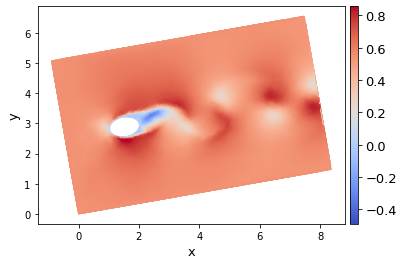}
\\
(d)  MultiScaleGNN & (e) MultiScaleGNNg & (f) MultiScaleGNNg
\\
train rots. (MAE = 0.0245) & train no rots. (MAE = 0.0997) & train rots. (MAE = 0.0587)
\end{tabular}
\caption{\small Target and predictions (after 100 time-steps) of the horizontal velocity field with a rotation of 10 degrees of the physical domain (sample from the validation dataset \texttt{NsVal} with $Re=864$).
}
\label{fig:rot10}
\end{figure}

\begin{figure}[ht]
\centering
\begin{tabular}{cccc}
\includegraphics[clip, height=0.15\textwidth, trim={15mm 15mm 25mm 5mm} ]{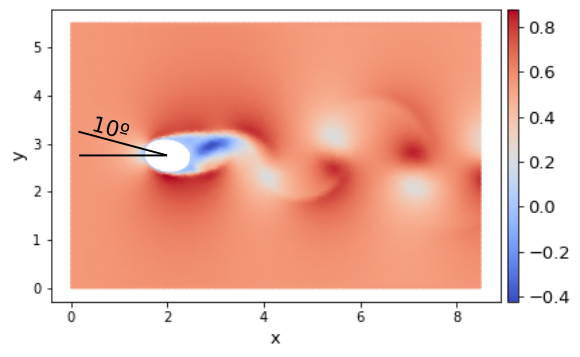} &
\includegraphics[clip, height=0.15\textwidth, trim={15mm 15mm 25mm 5mm}]{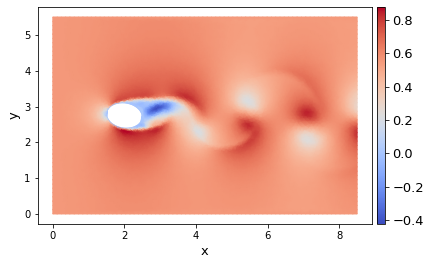} &
\includegraphics[clip, height=0.15\textwidth, trim={15mm 15mm 25mm 5mm}]{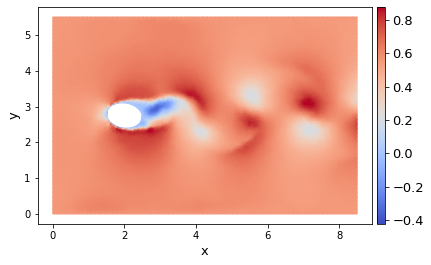} &
\includegraphics[clip, height=0.15\textwidth, trim={15mm 15mm 25mm 5mm}]{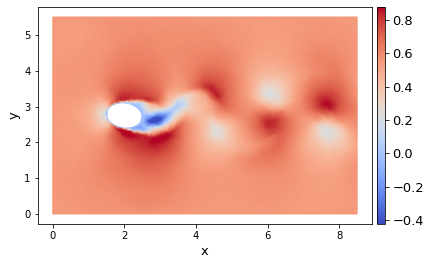}
\\
(a) Target & (b) \model{} & (c) MultiScaleGNN  & (d) MultiScaleGNNg
\\
& (MAE = 0.0148) & (MAE = 0.0223) & (MAE = 0.0735)
\end{tabular}
\caption{\small Target and predictions (after 100 time-steps) of the horizontal velocity field around an ellipse with an angle of attack of 10º and $Re=800$ (sample from the \texttt{NsAoA} dataset).
}
\label{fig:aoa}
\end{figure}

\model{} is rotation equivariant by design, whereas traditionally this symmetry is learnt through data augmentation of the training dataset with random rotations of the physical domain.
Figure \ref{fig:rot10} shows the ground truth and predictions after 100 time-steps of the horizontal velocity on a sample from the \texttt{NsVal} dataset that has been rotated 10 degrees.
It can be observed that the benchmark models trained without rotations produce unstable simulations, whereas all the other models produce realistic results.
Among these, \model{} has the lower MAE and produces a better resolved solution.
Figure \ref{fig:aoa} shows the ground truth and predictions after 100 time-steps of the horizontal velocity on a sample from the \texttt{NsAoA} dataset.
In this case the ellipse has been rotated 10 degrees clockwise (i.e. the angle of attack is 10 degrees).
It is possible to notice that in \model{}'s prediction the position of the wake and vortices, as well as their shape, is more similar to the ground truth than in the other two benchmark models (both trained with rotations).
This illustrates the better generalisation of \model{} with respect to models that are not designed to be rotation equivariant.
Simulations with the ground truth and predictions of \model{} and the benchmark models can be found \href{https://imperialcollegelondon.box.com/s/3zh4fwdh8tt7ifowlspwnb6uz2nwq5bs}{\textit{here.}}

% \begin{figure}[ht]
% \centering
% \subfloat[\centering Target]{\includegraphics[clip, width=0.3\textwidth, trim={15mm 15mm 25mm 5mm} ]{figures/NsVal350Rot20/Truth.png}}
% \quad
% \subfloat[\centering \model{} \\ MAE = 0.0210]{\includegraphics[clip, width=0.3\textwidth, trim={15mm 15mm 25mm 5mm}]{figures/NsVal350Rot20/Equi.png}}
% \quad
% \subfloat[\centering MultiScaleGNN \\ train no rots. \\ MAE = 0.1051]{\includegraphics[clip, width=0.3\textwidth, trim={15mm 15mm 25mm 5mm}]{figures/NsVal350Rot20/Classic.png}}
% \quad
% \subfloat[\centering MultiScaleGNN \\ train rots. \\ MAE = 0.0245]{\includegraphics[clip, width=0.3\textwidth, trim={15mm 15mm 25mm 5mm}]{figures/NsVal350Rot20/ClassicRot.png}}
% \quad
% \subfloat[\centering MultiScaleGNNg \\ train no rots. \\ MAE = 0.0997]{\includegraphics[clip, width=0.3\textwidth, trim={15mm 15mm 25mm 5mm}]{figures/NsVal350Rot20/Guillard.png}}
% \quad
% \subfloat[\centering MultiScaleGNNg \\ train rots. \\ MAE = 0.0587]{\includegraphics[clip, width=0.3\textwidth, trim={15mm 15mm 25mm 5mm}]{figures/NsVal350Rot20/GuillardRot.png}}
% \quad
% \caption{Target and predictions (after 100 time-steps) of the horizontal velocity field with a rotation of 20 degrees of the physical domain (sample from the validation dataset \texttt{NsVal} with $Re=864$).
% }
% \label{fig:rot20}
% \end{figure}

\end{document}